\documentclass[journal,twoside,web]{ieeecolor}
\usepackage{tmi}
\usepackage{cite}
\usepackage{amsmath,amssymb,amsfonts}
\usepackage{algorithmic}
\usepackage{graphicx}
\usepackage{textcomp}
\usepackage{booktabs} 
\usepackage{multirow} 
\usepackage{float}    
\usepackage{placeins}
\def\BibTeX{{\rm B\kern-.05em{\sc i\kern-.025em b}\kern-.08em
    T\kern-.1667em\lower.7ex\hbox{E}\kern-.125emX}}
\begin{document}
\title{A Learning-based Framework for Spatial Impulse Response Compensation in 3D Photoacoustic Computed Tomography}
\author{
    Kaiyi Yang, Seonyeong Park, Gangwon Jeong, Hsuan-Kai Huang, Alexander A. Oraevsky, \IEEEmembership{Member, IEEE}, Umberto Villa, \IEEEmembership{Member, IEEE}, and Mark A. Anastasio, \IEEEmembership{Fellow, IEEE}
    \thanks{This work was supported by NIH under Awards R01EB031585, P41EB031772 and R01EB034261.}
    \thanks{This work used the Delta system at the National Center for Supercomputing Applications through allocation MDE230007 from the Advanced Cyberinfrastructure Coordination Ecosystem: Services \& Support (ACCESS) program, which is supported by U.S. National Science Foundation grants \#2138259, \#2138286, \#2138307, \#2137603, and \#2138296.}
    \thanks{(Corresponding author: Mark A. Anastasio.) Kaiyi Yang and Hsuan-Kai Huang are with the Department of Electrical and Computer Engineering, University of Illinois Urbana–Champaign, Urbana, IL 61801, USA (e-mail: kaiyiy2@illinois.edu; hkhuang3@illinois.edu). Seonyeong Park, Gangwon Jeong, and Mark A. Anastasio are with the Department of Bioengineering, University of Illinois Urbana–Champaign, Urbana, IL 61801, USA (e-mail: sp33@illinois.edu; gangwon2@illinois.edu; maa@illinois.edu). Alexander A. Oraevsky is with TomoWave Laboratories, Inc., Houston, TX 77054, USA (e-mail: ao@tomowave.com). Umberto Villa is with the Oden Institute for Computational Engineering and Sciences and with the Department of Biomedical Engineering, The University of Texas at Austin, Austin, TX 78712, USA (e-mail: uvilla@austin.utexas.edu).}
}

\maketitle

\begin{abstract}

Photoacoustic computed tomography (PACT) is a promising imaging modality that combines the advantages of optical contrast with ultrasound detection. Utilizing ultrasound transducers with larger surface areas can improve detection sensitivity. However, when computationally efficient analytic reconstruction methods that neglect the spatial impulse responses (SIRs) of the transducer are employed, the spatial resolution of the reconstructed images will be compromised. Although optimization-based reconstruction methods can explicitly account for SIR effects, their computational cost is generally high, particularly in three-dimensional (3D) applications.
To address the need for accurate but rapid 3D PACT image reconstruction, this study presents a framework for establishing a learned SIR compensation method that operates in the data domain. The learned compensation method maps SIR-corrupted PACT measurement data to compensated data that would have been recorded by idealized point-like transducers.
Subsequently, the compensated data can be used with a computationally efficient reconstruction method that neglects SIR effects. Two variants of the learned compensation model are investigated that employ a U-Net model and a specifically designed, physics-inspired model, referred to as Deconv-Net. 
A fast and analytical training data generation procedure is also a component of the presented framework.
The framework is rigorously validated in virtual imaging studies, demonstrating resolution improvement and robustness to noise variations, object complexity, and sound speed heterogeneity. 
When applied to \textit{in-vivo} breast imaging data, the learned compensation models revealed fine structures that had been obscured by SIR-induced artifacts.
To our knowledge, this is the first demonstration of learned SIR compensation in 3D PACT imaging. 
\end{abstract}


\section{Introduction}
\label{sec:introduction}
\IEEEPARstart{P}{hotoacoustic} computed tomography (PACT), also known as optoacoustic tomography, is a hybrid imaging modality that combines the advantages of both optical imaging and ultrasound imaging~\cite{wang2012photoacoustic,wang2017photoacoustic}.
Unlike conventional ultrasound imaging, PACT provides rich optical contrast by detecting ultrasound waves generated from localized optical absorption. Pulsed laser irradiation induces the photoacoustic effect, producing an initial pressure distribution that reflects the tissue's optical energy deposition. 
The generated pressure field propagates through tissue and is detected by ultrasound transducers as acoustic signals that undergo significantly less scattering than light.
These advantages have contributed to its adoption in both preclinical and clinical imaging~{\cite{wang2009multiscale,wang2003noninvasive,andreev2003detection,kruger2010photoacoustic,ntziachristos2010molecular, oraevsky2018full}}.

Despite significant advancements in PACT, improving detection sensitivity remains an ongoing challenge. The minimum detectable pressure is limited by the thermal noise of the transducer~\cite{oraevsky2000ultimate}. Lowering this noise floor can improve the detection of weak signals originating from deeper tissues. For piezoelectric transducers, the root-mean-square thermal noise voltage is inversely proportional to the square root of the piezoelement's capacitance~\cite{andreev2003detection}. Increasing the transducer element's area is an effective approach to improve its capacitance, and consequently, the detection sensitivity~\cite{andreev2003detection, yang2007ring}. However, a larger transducer aperture also yields a narrower directivity pattern,
which is characterized by its spatial impulse response (SIR)~\cite{wang2010imaging}.

Many existing PACT reconstruction methods assume point-like transducers~\cite{xu2005universal,anastasio2005half}, neglecting the SIR.
If the SIR is not compensated for, the resulting reconstructed images will be degraded by a spatially varying blurring~\cite{wang2011bookchapter}.
Optimization-based iterative reconstruction methods~\cite{li2010model,wang2010imaging,rosenthal2011model} provide a principled approach to SIR compensation.
However, mitigating their high computational costs in three-dimensional (3D) applications remains challenging~\cite{wang2013accelerating}. 
Heuristic backprojection-based methods have been proposed \cite{luo2023influences,lu2020full} that are computationally efficient, but the reported results demonstrate only partial compensation for SIR-induced image degradation. Learning-based methods have shown great potential in various imaging problems~\cite{yanny2022deep,cam2024learning}, but their application to SIR compensation has thus far been limited to two-dimensional settings~\cite{munjal2024deep}.
There remains an important need for the further development and investigation of computationally efficient methods for SIR compensation in 3D PACT.

To address this need, a novel framework for learning-based SIR compensation is proposed, which consists of a data-domain SIR compensation model and a synthetic data generation strategy tailored for supervised training of the proposed model. The proposed SIR compensation model is employed to map PACT measurement data acquired by finite-size transducers to the corresponding data that would have been obtained by point-like transducers. Subsequently, reconstruction methods that assume point-like transducers can be adopted. Two variants of the SIR compensation model are investigated: one purely data-driven and another with a physics-inspired  design. Synthetic data are generated based on the configuration of the target PACT system and are sufficiently diverse to enable the trained models to perform well on experimental data. The proposed framework is quantitatively validated through virtual imaging studies that involve clinically relevant numerical breast phantoms (NBPs) and further evaluated by use of experimental \textit{in-vivo} breast PACT data. 
The present work significantly extends a preliminary study on this topic~\cite{yang2025compensating} by introducing an additional physics-inspired SIR compensation model, conducting systematic assessments, and providing experimental validation.


The remainder of the paper is organized as follows. A review of the salient background material is provided in Section~\ref{sect:background}. The proposed SIR compensation framework is detailed in Section~\ref{sect:method}. The framework is quantitatively validated via virtual imaging studies in Sections~\ref{sect:vit_studies} and \ref{sect:vit_results}. Section~\ref{sect:exp_study} presents experimental results from \textit{in-vivo} breast data. Finally, the article concludes with a summary of the key findings and a discussion of future directions in Section~\ref{sect:conclusion}.

\section{Background}
\label{sect:background}
\subsection{Continuous-to-Continuous (C-C) Imaging Model}
\label{sect:c-c}
In PACT, a short laser pulse irradiates an object, which absorbs the optical energy and generates a localized rise in acoustic pressure through the photoacoustic effect.
This sought-after laser-induced pressure rise, defined as the initial pressure distribution, is denoted as $p_0(\mathbf{r}):= p(\mathbf{r}, t=0)$, where $\mathbf{r} \in \mathbb{R}^3$ denotes spatial location, and $t$ represents time. The resulting pressure wavefield propagates through the medium and is detected by ultrasound transducers located at $\mathbf{r}_0\in\Omega_0\subset \mathbb{R}^3$, where $\Omega_0$ denotes the measurement aperture whose convex hull fully encloses the object. Under the assumptions of instantaneous optical illumination and an acoustically homogeneous, lossless medium, with uniform speed of sound (SOS) $c_0$ matching that of the acoustic coupling medium, the pressure detected by an ideal point-like transducer is expressed by the C-C imaging model~\cite{wang2017photoacoustic}:
\begin{equation}
    p^{\text{ideal}}(\mathbf{r}_0, t) = \frac{1}{4\pi c_0^2} \int_{V} d^3\mathbf{r} \, p_0(\mathbf{r}) \,  h^{\text{ideal}}(\mathbf{r}_{0}, \mathbf{r}, t),
    \label{eq:forward_model}
\end{equation}
where the point response function  $h^{\text{ideal}}(\mathbf{r}_{0}, \mathbf{r}, t)\equiv \frac{d}{dt} \frac{\delta \left( t - \frac{|\mathbf{r}_0 - \mathbf{r}|}{c_0} \right)}{|\mathbf{r}_0 - \mathbf{r}|}.$
Here, $V$ represents the object's support and $\delta(t)$ is the temporal Dirac delta function.
\subsection{Continuous-to-Discrete (C-D) Imaging Model with SIR}
\label{sect:c-d}
In practice, the photoacoustic pressure field is degraded by the response of the ultrasonic transducers and sampled during the measurement process.
Consider that measurements are acquired at $N_q$ transducer locations. At each transducer location $\mathbf{r}_{0,q}$, $N_t$ temporal pressure samples are collected at time intervals of $\Delta t$. Assuming ideal, point-like transducers, let the vector $\textbf{p}^{\text{ideal}} \in \mathbb{R}^M$ denote the lexicographically ordered sampled data, with $M = N_q N_t$ total samples. The $r$-th temporal sample recorded at the $q$-th transducer is written as: 
\begin{equation}
[\textbf{p}\!^{\text{ideal}}]_{qN_t+r}= p^{\text{ideal}}({\mathbf{r}_{0,q}}, t) \big|_{t=r\Delta t}, 
    \begin{array}{c}
    \scriptstyle r = 0, \ldots, N_t-1 \\[-0.4em]
    \scriptstyle q = 0, \ldots, N_q-1
    \end{array}.
    \label{eq:C-D_forward_model}
\end{equation}

When the ultrasound transducer has a finite aperture, the recorded acoustic pressure corresponds to a spatial average of the pressure field over the transducer's aperture. Let the vector $\textbf{p}^{\text{finite}} \in \mathbb{R}^M$ denote the discrete measurement data acquired by use of finite-sized transducers~\cite{wang2011bookchapter,wang2017photoacoustic}:
\begin{equation}
[\textbf{p}^{\text{finite}}]_{qN_t+r}
    = \frac{1}{S_q} \int_{S_q} d\mathbf{r}_0' \, p^{\text{ideal}}(\mathbf{r}_0', t) \bigg|_{t=r\Delta t},
\label{eq:surface_integral}
\end{equation}
where $S_q$ denotes the active aperture area of the $q$-th transducer centered at $\mathbf{r}_{0,q}$.
For each point acoustic source located at $\mathbf{r}$, the surface integral in \eqref{eq:surface_integral} can be approximated in the form of one-dimensional (1D) convolution~\cite{wang2010imaging}. The C-D imaging model with consideration of the SIR can be expressed as:
\begin{align}
\label{eq:forward_model_finite_transducers}
[\textbf{p}^{\text{finite}}]_{qN_t+r}
= \int_V d^3\mathbf{r} \, &p_0(\mathbf{r})\,
\big[h^{\text{ideal}}(\mathbf{r}_{0,q},\mathbf{r},t) \\[-0.25em]
&*_t h^{\text{SIR}}(\mathbf{r}_{0,q},\mathbf{r},t)\big]\,
\nonumber \big|_{t=r\Delta t},
\end{align}
where $*_t$ represents 1D temporal convolution and $h^{\text{SIR}}(\mathbf{r}_{0,q}, \mathbf{r}, t)$ denotes the SIR of the $q$-th transducer.

Analytic expressions for the SIR have been derived for several canonical transducer aperture shapes~\cite{stepanishen1971transient,jensen1999new}. 
For example, under far-field assumptions, the SIR of a flat rectangular transducer with dimensions $a\times b$ can be expressed in the temporal frequency domain as~\cite{MITSUHASHI201421}: 
\begin{equation}{\parbox{0.9\linewidth}{\centering 
\begin{math}
\tilde{h}^{\text{SIR}}(\mathbf{r}_{0,q}, \mathbf{r}, f) = \text{sinc}\left(\frac{a {x}^\text{local}_q \pi f}{c_0 |\mathbf{r}_{0,q} - \mathbf{r} |} \right) \text{sinc}\left(\frac{b {y}^\text{local}_q \pi f}{c_0 |\mathbf{r}_{0,q} - \mathbf{r} |} \right). 
\end{math}}}
\label{eq:rect_transducer_sir}
\end{equation}
Here, $f$ is the temporal frequency variable conjugate to $t$, and a `$\sim$' above a function denotes its Fourier transform. The coordinates $({x}^\text{local}_q,{y}^\text{local}_q, {z}^\text{local}_q)$ represent the position of a point source in a local coordinate system of the $q$-th transducer.
Notably, the SIR varies spatially depending on 
the source position relative to the transducer center. 

A discrete-to-discrete (D-D) imaging model that incorporates the SIR can be derived from the C-D model by approximating the continuous object $p_0(\mathbf{r})$ as a linear combination of a finite set of expansion functions, such as radially symmetric blobs~\cite{wang2010imaging,wang2014discrete}. While such models support optimization-based image reconstruction and data synthesis, they inherently introduce discretization errors and often impose substantial computational burdens, even when leveraging parallel computing on GPUs~\cite{wang2013accelerating}.

\section{Learning-based Framework for Data Space SIR Compensation}
\label{sect:method}
\normalcolor
\begin{figure}
    \centering
    \includegraphics[width=3.5in]{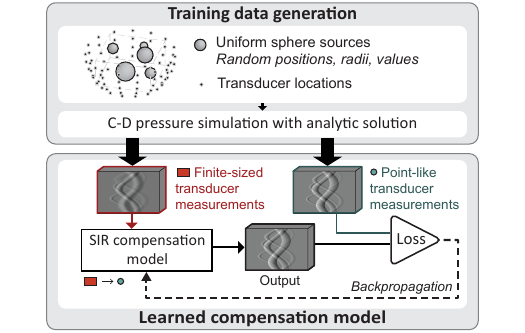}
    \caption{
    Learning-based framework for data space SIR compensation, consisting of a data generation approach (top) tailored for training of an SIR compensation model (bottom). 
    The SIR compensation model is trained to approximate the mapping from pressure data acquired with finite-sized transducers (input, marked in red) to their point-like equivalents (target, marked in green). 
    Training data were generated efficiently using the analytical C-D forward formula of sphere sources with stochastically sampled positions, radii, and initial pressures.
    }
    \label{fig:method_overview}
\end{figure}

A learning-based framework is proposed for SIR compensation in the measurement domain.
As illustrated in Fig.~\ref{fig:method_overview}, the framework consists of a learned SIR compensation model that approximates the mapping from pressure data acquired with finite-sized transducers (network inputs) to the corresponding point-like transducer measurements (network targets), along with a data generation strategy tailored for supervised training. 

In this study, a rectangular transducer detection surface is considered, motivated by an existing 3D PACT system~\cite{oraevsky2018full}. 
However, the proposed framework is applicable to any transducer surface shape and array geometry for which the SIR can be numerically approximated~\cite{jensen1999new}. 
Once trained, the learned SIR compensation model can be employed in a two-step image reconstruction procedure, in which the SIR-compensated pressure data serve as input to a computationally efficient 
reconstruction method that assumes point-like transducers. 

The training data generation strategy is described in Section~\ref{sec:framework_for_simulating_training_data}. 
Two formulations of the data space SIR compensation model are considered: a purely data-driven approach, described in Section~\ref{sec:end_to_end_learning_approach}, and a physics-inspired approach, detailed in Section~\ref{sec:physics_inspired_learning_approach}.

\subsection{Stochastic Spheres Data for Model Training}
\label{sec:framework_for_simulating_training_data}
A synthetic data generation strategy is proposed to enable supervised training of the SIR compensation models. 
Synthetic paired measurement data corresponding to the use of finite-sized transducers (inputs) and their point-like counterparts (targets) were computed. This is necessary because acquiring such paired experimental data is generally infeasible, as, in practice, transducer elements with a very small active surface area have low signal-to-noise ratio. 
The proposed data generation procedure yields system-specific training data. Importantly, as demonstrated later, the resulting training data are sufficiently diverse to enable the trained compensation models to perform well when applied to experimental measurements.


To achieve this, collections of continuous uniform spherical sources with stochastically assigned radii, positions, and initial pressure amplitudes can be used to construct synthetic to-be-imaged initial pressure distributions (referred to as "objects"). This design offers several key advantages. First, the superposition of stochastic spheres enables the construction of complex and heterogeneous objects without imposing a particular structural prior. Second, the random placement of spheres relative to the transducer array results in a wide range of SIRs in the simulated measurements. 
Collectively, these factors enhance training-data diversity to support model generalization.
Additionally, the use of uniform spherical sources enables closed-form computation of measurements via the C-D imaging model~\cite{wang2017photoacoustic, wang2011bookchapter}, thereby bypassing the need for object discretization required in the D-D imaging model. This eliminates object discretization errors and improves computational efficiency by computing measurements per sphere rather than per grid point~\cite{wang2010imaging}. Because the number of spheres is typically much smaller than the number of voxels, this strategy reduces the computational cost compared to the generally expensive D-D imaging model, as discussed in Section~\ref{sect:c-d}.

In practice, each synthetic object consists of multiple, potentially overlapping spheres whose centers are stochastically sampled within the support of the to-be-imaged objects.
The sphere radii and initial pressures are drawn from biologically plausible ranges that reflect typical tissue structures in PACT imaging.
Using these objects, synthetic input-target measurement pairs, $\{\mathbf{p}^{\text{rect}}, \mathbf{p}^{\text{ideal}}\}$, are computed using closed-form models: $\mathbf{p}^{\text{rect}}$ via \eqref{eq:forward_model_finite_transducers} and \eqref{eq:rect_transducer_sir}, and $\mathbf{p}^{\text{ideal}}$ via \eqref{eq:forward_model} and \eqref{eq:C-D_forward_model}, assuming a nominal SOS value. The resulting dataset is hereafter referred to as the Stochastic Spheres dataset.

\subsection{Purely Data-Driven Compensation Model} 
\label{sec:end_to_end_learning_approach}
In the purely data-driven approach, the SIR compensation model is formulated as a learned operator that directly maps the full set of pressure time traces recorded by finite-sized transducers to their point-like counterparts.
The SIR is dependent on the relative position between the acoustic source and the transducer recording the pressure. As described in \eqref{eq:forward_model} and \eqref{eq:forward_model_finite_transducers}, each recorded time trace integrates pressure waves originating from all source locations within a given support, with arrival times reflecting their distances from the transducer. While this temporal information allows estimation of source distance, directional information can only be recovered by considering time traces recorded at multiple transducer locations. 
Therefore, a single time trace is insufficient for accurate SIR estimation and compensation, motivating the use of full-array pressure data in the input and target of the model. 
\begin{figure*}[t]
    \centering
    \includegraphics[width=7.16in]{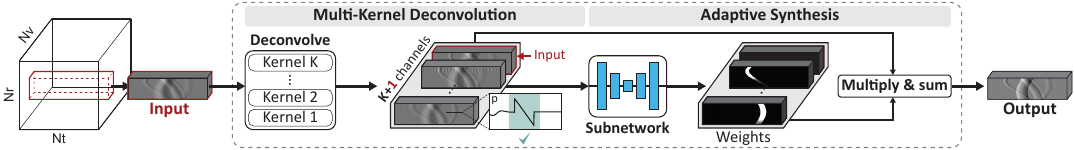} 
    \caption{Illustration of the Deconv-Net architecture for SIR compensation. The leftmost 3D schematic represents the full-array pressure measurements, with the red-outlined region indicating a cropped subset used as the network input.  
    This input is first processed by a multi-kernel 1D deconvolution module, where $K$ learned kernels model the spatially variant SIR at different relative positions between the source and transducer. 
    When a deconvolved kernel closely approximates the underlying SIR, the resulting signal contains locally compensated segments (exemplified by the green-highlighted region).  In the subsequent adaptive synthesis step, the $K$ deconvolved signals, together with the input signal, are concatenated and passed through a subnetwork that estimates a weight tensor, which is then used to combine them via element-wise multiplication and channel-wise summation, producing the final SIR-compensated output.  
    }
    \label{fig:deconvnet}
    \vspace{-0.2cm}
\end{figure*}
To learn a mapping from pressure measurements acquired with finite-sized transducers to their point-like transducer equivalents, a 3D U-Net architecture~\cite{cciccek20163d} can be adopted. 
The 3D architecture can be specified to accommodate the structure of the measurement data, which are reshaped from vectors $\mathbf{p}^{\text{rect}}, \mathbf{p}^{\text{ideal}} \in \mathbb{R}^{M}$ into tensors $\mathbf{P}^{\text{rect}}, \mathbf{P}^{\text{ideal}} \in \mathbb{R}^{N_r\times N_v\times N_t}$, where $M = N_q N_t = N_r N_v N_t$. Here, $N_r$ represents the number of transducer elements per tomographic view, and $N_v$ denotes the total number of tomographic views. 
This tensor representation allows the network to leverage spatial and temporal structures inherent in the data.

\subsection{Physics-Inspired Learned Compensation Model}
\label{sec:physics_inspired_learning_approach}
The physics-inspired approach builds on empirical evidence that problem-specific neural networks 
incorporating physics knowledge 
can outperform purely data-driven models~\cite{banerjee2024physics}.
Accordingly, the proposed Deconv-Net integrates learnable deconvolution filters that embed the physical SIR model within a data-driven framework, inspired by recent advances in learned inverse problem solvers~\cite{yanny2022deep, li2024tabe}.

As described in \eqref{eq:forward_model_finite_transducers}, when finite-aperture detectors are deployed, the recorded pressure can be modeled as a non-stationary temporal convolution of the ideal pressure signal with the spatially variant SIR.
Conceptually, this process can be computationally undone by solving the corresponding inverse problem.
Deconv-Net performs SIR compensation in two steps, as illustrated in Fig.~\ref{fig:deconvnet}.
First, the SIR-degraded input signals are deconvolved using a set of learned kernels approximating the SIRs at different locations.
At spatiotemporal regions where the kernel accurately represents the underlying SIR, the deconvolution yields locally compensated intermediate signal segments.
Second, a subnetwork estimates weights that combine the intermediate signals into a single compensated output, approximating the measurement that would have been acquired by point-like transducers.
Deconv-Net is implemented as a single, modular network that integrates both steps, trained in an end-to-end manner. 
Each step is detailed in the following subsections. 
%
\subsubsection{Deconvolution of SIR with Learnable Parameters} 
This step performs $K$ Fourier-space deconvolutions, each with a kernel approximating the SIR corresponding to a distinct spatial location, denoted as ${h}_k^{\text{SIR}}$ for $k=1, \cdots, K$. While using a large number of kernels $K$ enables more detailed modeling of the spatially varying SIR, it is accompanied by an increase in GPU memory usage. 
To accommodate more kernels while reducing memory requirements, the input and target to Deconv-Net are pressure signals recorded using a subset of the transducer array elements, referred to as a ``patch'', rather than the full-array data, as illustrated in Fig.~\ref{fig:deconvnet}.
Denoting a patch by $\mathcal{P}$, the input and target tensors are $\mathbf{P}^{\text{rect}}_\mathcal{P}, \mathbf{P}^{\text{ideal}}_\mathcal{P}\in \mathbb{R}^{N_r^\mathcal{P}\times N_v^\mathcal{P}\times N_t}$, respectively.
Each patch consists of the full-length pressure time traces of size $N_t$, recorded by a subarray of transducers spanning $N_r^\mathcal{P}$ neighboring elements on the probe and $N_v^\mathcal{P}$ adjacent tomographic views. 

The deconvolution can be implemented via a differentiable 1D Wiener deconvolution layer, adapted from~\cite{yanny2022deep}, and applied to each time trace within the input patch $\mathbf{P}_{\mathcal{P}}^{\text{rect}}$. 
Deconvolving $K$ kernels yields the four-dimensional tensor $\mathbf{P}^{\text{deconv}}_\mathcal{P} \in \mathbb{R}^{K\times N_r^\mathcal{P} \times N_v^\mathcal{P} \times N_t}$. Let $[\mathbf{P}^{\text{rect}}_{\mathcal{P}}]_{(i_r, i_v)}\in \mathbb{R}^{N_t}$ denote the 1D pressure time trace extracted from the 3D tensor $\mathbf{P}^{\text{rect}}_\mathcal{P}$, which is associated to the $(i_r, i_v)$-th transducer in the subarray. The corresponding deconvolved signal with the $k$-th kernel is: 
\begin{align}\label{eq:wiener_deconvolution}
    \nonumber[\mathbf{P}_{\mathcal{P}}^{\text{deconv}}]_{(k, i_r, i_v)} &=\\[-0.10em]\mathcal{F}_{1D}^{-1}&\Biggl\{\frac{{\tilde{h}_k^{\text{SIR}}(f_l)}^{*} \mathcal{F}_{1D}\{{\mathbf{P}}_{(i_r, i_v)}^{\text{rect}}\}}{|\tilde{h}_k^{\text{SIR}}(f_l)|^2 + \lambda_k} \Biggl\} \Biggl|_{f_l=l\Delta f}. 
\end{align}
Here, $\mathcal{F}_{1D}$ and $\mathcal{F}_{1D}^{-1}$ denote the 1D discrete Fourier transform and its inverse, respectively. The deconvolution kernel, $\tilde{h}_k^{\text{SIR}}(f_l)$, represents the Fourier domain expression of the $k$-th SIR kernel, sampled at discrete frequencies $f_l = l\Delta f$ with an interval $\Delta f$. The learnable regularization parameter $\lambda_k$ associated with the $k$-th kernel controls the balance between noise suppression and signal preservation~\cite{yanny2022deep}.  

In \eqref{eq:wiener_deconvolution}, each kernel ${h}_k^{\text{SIR}}$ seeks to emulate the spatially varying SIR associated with a certain relative position between the source and detector.
Leveraging the closed-form SIR expression \eqref{eq:rect_transducer_sir} of rectangular transducers, each SIR kernel can be modeled by a parameterized function with a small number of learnable parameters.
Given the known imaging geometry, the transducer dimensions $a$ and $b$ can be set a priori. The SOS $c_0$ is typically set to a nominal value of $1.5~\mathrm{mm}/\mu\mathrm{s}$.
The spatial dependence of the SIR is fully captured by the local coordinates ${\textbf{r}}^\text{local}_k=({x}^{\text{local}}_k, {y}^\text{local}_k, {z}^\text{local}_k)$, 
which specify the source position with respect to the local reference system of each transducer. 
The $K$ source locations ${\textbf{r}}^\text{local}_k$ and their respective regularization parameters $\lambda_k$ for $k=1,\ldots,K$, are learned during training. 

\subsubsection{Adaptive Synthesis of Deconvolved Signals}
To produce the SIR-compensated signals, a weighted summation is performed 
over the $K$ deconvolved signals $\mathbf{P}_\mathcal{P}^{\text{deconv}}$ from the previous step and 
the original non-deconvolved input $\mathbf{P}_\mathcal{P}^{\text{rect}}$. 
These $K+1$ components are concatenated into a joint tensor $\mathbf{P}^{\text{join}}_\mathcal{P}\in\mathbb{R}^{(K+1)\times N_r^\mathcal{P}\times N_v^\mathcal{P}\times N_t}$, which serves as input to a weight-learning subnetwork. At each spatiotemporal location $(i_r, i_v, r)$, the network computes a set of weights via a channel-wise softmax, yielding a weighting tensor $\mathbf{W} \in \mathbb{R}^{(K+1)\times N_r^\mathcal{P} \times N_v^\mathcal{P} \times N_t}$. Each weight satisfies $0 \le \mathbf{W}_{(k, i_r, i_v, r)} \le 1$ and $\sum_{k=1}^{K+1} \mathbf{W}_{(k, i_r, i_v, r)} = 1$, with higher values assigned to components that better approximate point-like transducer measurements. 
The final SIR-compensated output \( \mathbf{P}^{\text{comp}}_{\mathcal{P}} \in \mathbb{R}^{N_r^\mathcal{P}\times N_v^\mathcal{P}\times N_t}\) is computed as a weighted summation at each spatiotemporal index: 
\begin{equation}
\mathbf{P}^{\text{comp}}_{\mathcal{P}(i_r, i_v, r)} = \sum_{k=1}^{K+1} \mathbf{W}_{(k, i_r, i_v, r)}\cdot \mathbf{P}^{\text{join}}_{\mathcal{P}(k, i_r, i_v, r)}.
\end{equation}

The Deconv-Net is trained end-to-end using input-target patch pairs $\mathbf{P}^{\text{rect}}_\mathcal{P}$ and $\mathbf{P}^{\text{ideal}}_\mathcal{P}$, extracted at random from $\mathbf{P}^{\text{rect}}$ and $\mathbf{P}^{\text{ideal}}$, respectively. 
The extraction accounts for the periodicity of the tomographic view angles, treating the first and last views as adjacent. During inference, the full-array data $\mathbf{P}^{\text{rect}}$ are processed by applying Deconv-Net to each patch in a sliding-window manner to obtain the full compensated result $\mathbf{P}^{\text{comp}}$.

\section{Virtual Imaging Studies} \label{sect:vit_studies}
\normalcolor
This section outlines the virtual imaging studies conducted to quantitatively assess the proposed SIR compensation framework. 
The trained networks were first validated on an in-distribution (ID) test set, which comprised a held-out subset of the Stochastic Spheres dataset used for training.
To further examine generalizability, out-of-distribution (OOD) test sets were constructed, each designed for specific objectives: (i) systematic analysis of image resolution at varying distances from the measurement aperture, (ii) evaluation of robustness to structurally complex objects with anatomically and optically realistic features, and (iii) further assessment of robustness under additional realistic acoustic heterogeneity. 

\subsection{Virtual Imaging System Configuration}
\label{sect:virtual_imaging_system_configuration}
A virtual imaging system was configured based on 
an existing PACT breast imager, LOUISA-3D~\cite{oraevsky2018full}. 
The optical illumination subsystem was modeled as broadening slit light sources (half-angle of 12.5$^\circ$). 
The acoustic measurement subsystem employed a hemispherical measurement aperture with a radius of 85 mm. 
A transducer arc was positioned on this surface, spanning polar angles 
from $90.25^{\circ}$ to $170.25^{\circ}$, 
along which $96$ ultrasound transducers were equi-angularly distributed. 
The arc was rotated in $320$ equally spaced angular increments around a full $360^\circ$ circle. Each transducer acquired $2267$ time samples at a 20 MHz sampling rate. 
Rectangular transducers measuring 1.2 mm $\times$ 6 mm were deployed, 
with their long axis oriented along the tomographic view direction.

\subsection{Virtual PACT Imaging and Data Acquisition}
\label{sect:virtual_pact_imaging_and_data_acquisition}
Based on the virtual imaging system described above, four datasets were generated, each consisting of paired pressure data obtained using rectangular transducers ($\mathbf{P}^\text{rect}$) and their point-like counterparts ($\mathbf{P}^\text{point}$). One dataset was used for model training and validation, and three datasets were employed for OOD testing. These consist of: (i) the Stochastic Spheres dataset, defined in Section~\ref{sec:framework_for_simulating_training_data}; (ii) a dataset produced from six identical spheres at fixed positions, referred to as the Deterministic Spheres dataset; and (iii) two datasets generated employing stochastic 3D NBPs~\cite{park2023stochastic}, referred to as the NBP-Homogenous and NBP-Heterogeneous datasets, depending on the assumed SOS distribution. For all datasets, a lossless medium was assumed in the acoustic simulations.
\subsubsection{Stochastic Spheres Dataset}\label{sect:stochastic_spheres_dataset}
The definition of the imaged objects for this dataset was introduced in Section~\ref{sec:framework_for_simulating_training_data}. 
The initial pressure distributions were directly assigned through stochastic sampling without simulating the optical forward process, whereas the acoustic data acquisition was simulated following the system configuration described in Section~\ref{sect:virtual_imaging_system_configuration}.
Each object consisted of 200 uniform spheres. Sphere radii were sampled from a truncated log-normal distribution with a mean of 0.375 mm (representing typical human blood vessel thickness) and a standard deviation of 1 mm, constrained between 0.125 mm and 5 mm to reflect biologically relevant structures in PACT breast imaging~\cite{park2023stochastic}. Sphere positions were randomly sampled from a uniform distribution supported within a 60 mm-radius hemispherical field-of-view, considering the breast stabilization cup in LOUISA-3D~\cite{oraevsky2018full}. Initial pressure values were drawn from a uniform distribution ranging from 0 to 0.02 arbitrary units (AU).
Pressure data were computed following the procedure described in Section~\ref{sec:framework_for_simulating_training_data}, assuming a homogeneous medium with a nominal speed of sound of 1.5~mm/$\mu$s.
Independent and identically distributed (i.i.d.) Gaussian noise with zero mean was added to the simulated pressure data. 
The noise standard deviation was set to 2.67\% of the 90th percentile of the
signal power in the noiseless $\mathbf{P}^{\text{rect}}$ data, empirically estimated from experimental data (Section~\ref{sect:exp_study}). This dataset consists of $10,000$ pairs of measurements, split into 70\% training, 10\% validation, and 20\% testing subsets. 

\subsubsection{Deterministic Spheres Dataset}\label{sect:deterministic_spheres_dataset}
This dataset, specifically designed for analyzing spatial resolution, was generated from an object consisting of six identical spheres with a radius of 1.2 mm embedded in an optically non-absorbing background. As described in Section~\ref{sect:virtual_imaging_system_configuration}, the hemispherical measurement aperture was centered at the origin and spans the lower ($z<0$) hemisphere. 
The sphere centers were positioned at $(x, y, z)=(10n-5, 0, -2)$ mm for $n=1,\dots,6$. An initial pressure value of 1 AU was assigned to each sphere without simulating the optical forward process. 
Acoustic data acquisition was simulated in the same way as described above for the Stochastic Spheres dataset, with additional variations in noise level and background SOS. 
Four subsets of the Deterministic Spheres dataset were created: 
(i) baseline assumes the same noise level (2.67\%) and SOS ($c_0=1.5$ mm/$\mu$) as the Stochastic Spheres dataset (Section ~\ref{sect:stochastic_spheres_dataset}), which was used to train the SIR Compensation models; 
(ii) high noise assumes 10$\times$ baseline noise level; (iii) low SOS assumes $c_0=1.447$ mm/$\mu$s; (iv) high SOS assumes $c_0=1.555$ mm/$\mu$s. The two additional SOS values alongside the baseline correspond to the lower and upper bounds of the reported range for breast tissue SOS~\cite{chen2025learning}.

\normalcolor
\subsubsection{Stochastic NBP Datasets}
\normalcolor
Two OOD datasets were produced using 3D NBPs, reflecting anatomical realism and optical contrast derived from simulated photon transport. Additionally, one of the datasets incorporates heterogeneous SOS distributions to account for acoustic variability.
A total of 80 NBPs were stochastically generated following the modeling framework proposed by Park \textit{et al.}~\cite{park2023stochastic}, with 20 phantoms corresponding to each of 
the four BI-RADS categories
~\cite{american2018acr} of breast density: 
almost entirely fatty, scattered fibroglandular, heterogeneously dense, and extremely dense. 
Based on the system configuration described in Section~\ref{sect:virtual_imaging_system_configuration}, optical fluence in  
each phantom was simulated using 
the \texttt{MCX} software\cite{fang2009monte} on a 680$\times$680$\times$340 
grid with an isotropic voxel size of 0.25~mm. 
The resulting fluence was then used to generate the initial pressure distribution on the same grid. 
Details of the simulation are available in~\cite{park2023stochastic}. 

Photoacoustic wave propagation and data acquisition were modeled by the first-order acoustic wave equation~\cite{tabei2002k}, discretized and solved using the $k$-space pseudospectral time-domain method~\cite{treeby2010photoacoustic}. A validated implementation leveraging GPU acceleration capability of the Python package \texttt{JAX}~\cite{jax2018github} was employed, inspired by the software \texttt{j-Wave} \cite{stanziola2023j}. The acoustic medium was defined with a fixed ambient density of 1~g/cm$^3$ for all phantoms. For the NBP-Homogeneous dataset, a uniform SOS of 1.5~mm/$\mu$s was assumed. In contrast, the NBP-Heterogeneous dataset was simulated employing phantom-specific SOS distributions, with values assigned independently to each tissue type in each phantom, 
following~\cite{park2023stochastic}.
In each paired dataset ($\mathbf{P}^{\text{rect}}$, $\mathbf{P}^{\text{point}}$), the pressure data corresponding to rectangular transducers ($\mathbf{P}^{\text{rect}}$) were computed by spatially averaging the pressure field over each sensor surface, following the method described by Wise \textit{et al.}~\cite{wise2019representing}. 
Measurements were corrupted with i.i.d. Gaussian noise as 
described in Section~\ref{sect:stochastic_spheres_dataset}, with the standard deviation computed separately for the NBP-Homogeneous and NBP-Heterogeneous datasets.



\subsection{Model Implementation and Training}
\label{sect:model_training_and_image_recon}
The implemented purely data-driven SIR compensation model retained the structure
of the original 3D U-Net~\cite{cciccek20163d}, but included some adaptations. 
Cyclic padding was deployed in the view dimension within each convolution block to preserve spatial continuity between the first and last views. 
Max-pooling and up-sampling layers were replaced with strided convolution layers, and batch normalization was applied after each convolutional layer to stabilize training. 
In the physics-inspired SIR compensation model, the input patch size and the number of kernels were set to $N_r^\mathcal{P} = N_v^\mathcal{P} = 32$ and $K=127$, respectively.
The subnetwork employed to compute synthesis weights is a 3D U-Net.

The models were trained using the mean absolute error (MAE) loss and optimized with Adam (learning rate: $3 \times 10^{-4}$). 
The learning rate was halved if the validation loss did not improve for two consecutive epochs. Training was stopped after 
five consecutive epochs without improvement. 
Training was performed on two NVIDIA A40 GPUs with a batch size of 2.
When evaluated on the testing subset of the Stochastic Spheres dataset, the trained networks were able to reduce the objective MAE loss without overfitting to the training set.

\subsection{Image Reconstruction}
The restored pressure data output by the model can be utilized by any image reconstruction method that assumes point-like transducers.
In this study, the universal backprojection (UBP) method \cite{xu2005universal} was selected due to its widespread use and computational efficiency. 
For acoustically homogeneous data, the implementation of UBP assumed the true SOS distribution. For the NBP-Heterogeneous dataset, a two-region UBP method was employed, where the average true SOS in the breast and water regions were used.
The dimensions of the reconstructed images were 
$480\times 480\times 240$ voxels, with a voxel size of $0.25$ mm, corresponding to a field-of-view of 120 mm $\times$ 120 mm $\times$ 60 mm, matching 
the assumed breast stabilization cup. 

\begin{figure*}[!t]
    \centering
    \includegraphics[width=7.16in]{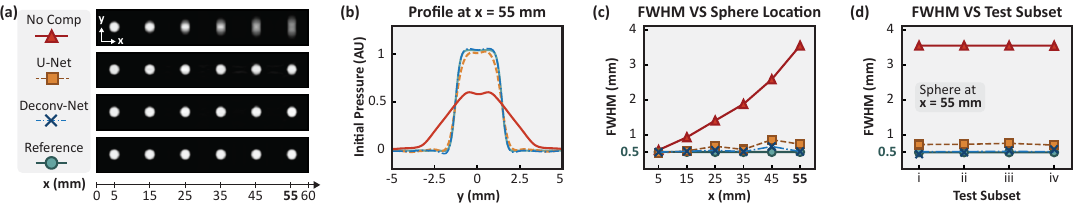}
    \caption{Results from Virtual Imaging Study 1. (a) Cross-sections at $z = -2$ mm under ID conditions: ${\mathbf{P}}^{\text{rect}}_0$, ${\mathbf{P}}^\text{U-Net}_0$, ${\mathbf{P}}_0^\text{Deconv-Net}$, and ${\mathbf{P}}_0^{\text{point}}$ (top to bottom). (b) Line profiles at $x = 55$ mm, with colors following the legend in (a). (c) Estimated FWHM across all sphere locations. (d) FWHM at $x = 55$ mm across all test subsets. Compensation by U-Net (orange dashed) and Deconv-Net (blue dashed) yields resolution closely matching the reference (turquoise solid) across all test subsets, as described in Section~\ref{sect:deterministic_spheres_dataset}: (i) baseline; (ii) high noise; (iii) low SOS; (iv) high SOS.}
    \label{fig:study_1}
\end{figure*}

\subsection{Study Designs and Evaluation}
\label{sect:study_designs_and_evaluation}
The performances of the learned SIR compensation models, U-Net (Section~\ref{sec:end_to_end_learning_approach}) and Deconv-Net (Section~\ref{sec:physics_inspired_learning_approach}), were evaluated in three studies.
Given that SIR induces spatially varying blurring, 
Study 1 was designed to evaluate spatial resolution in a controlled setting. To further examine model robustness under increasingly realistic conditions, Studies 2 and 3 incorporated anatomical and optical realism, with Study 3 additionally accounting for SOS heterogeneity. In all studies, pressure data simulated with rectangular transducers ($\mathbf{P}^\text{rect}$) were used as input to the learned models, producing SIR-compensated outputs $\mathbf{P}^\text{U-Net}$ and $\mathbf{P}^\text{Deconv-Net}$. The images reconstructed from these outputs are denoted as ${\mathbf{P}}_0^\text{U-Net}$ and ${\mathbf{P}}_0^\text{Deconv-Net}$, respectively.
These were compared against images reconstructed from $\mathbf{P}^\text{rect}$ without SIR compensation, denoted as ${\mathbf{P}}_0^\text{rect}$, and the reference images reconstructed from noiseless point-like transducer data $\mathbf{P}^\text{point}$, denoted as ${\mathbf{P}}_0^\text{point}$.


In Study 1, the Deterministic Spheres dataset, specifically designed for spatial resolution analysis, was employed. The proposed models were evaluated under two different noise levels and three homogeneous SOS conditions, as detailed in Section~\ref{sect:deterministic_spheres_dataset}. 
Spatial resolution was quantified using full width at half maximum (FWHM), and reconstructed object profiles were compared. Following the resolution analysis by Mitsuhashi \textit{et al.}~\cite{MITSUHASHI201421}, Gaussian smoothing was applied to the measurement data prior to image reconstruction to yield a theoretical FWHM lower bound of 0.5 mm.
FWHM was computed by fitting the y-profiles of each reconstructed sphere to a Gaussian-blurred rectangular model, as described in \cite{MITSUHASHI201421}.

Studies 2 and 3 were conducted 
using the NBP-Homogeneous and NBP-Heterogeneous datasets, respectively. Evaluation metrics included relative squared error (RSE)~\cite{park2023stochastic}, normalized cross-correlation (NCC), and the DICE coefficient. These metrics were computed within two concentric hemispherical shells at distances of 25–35 mm (superficial) and 35–45 mm (deep) from the measurement surface. These regions were chosen to capture the spatially varying effects of SIR near the hemispherical measurement aperture while avoiding potential bias from low-contrast regions due to optical attenuation at greater depths. 
To compute the DICE coefficient, vessel maps were extracted from reconstructed images. 
Although true optical fluence maps are not accessible in practice, they were used to fluence-compensate each reconstructed image via element-wise division, isolating the effect of SIR compensation on vessel structure preservation. 
The resulting images were then filtered with a multi-scale Frangi vesselness filter (\(\sigma = 1\)–5 voxels)~\cite{frangi1998multiscale} and thresholded using the triangle algorithm~\cite{zack1977automatic}. Voxels above the threshold were labeled as vessels, and the corresponding maps extracted from \({\mathbf{P}}_0^\text{point}\) were used as the reference. 
Statistical significance was evaluated via one-sided Mann–Whitney U tests~\cite{mann1947test}, testing whether each compensation method produced strictly lower RSE or higher NCC and DICE than its comparator. Differences yielding \(p< 0.05\) were considered significant.

\begin{figure*}[!ht] 
    \centering    \includegraphics[width=7.16in]{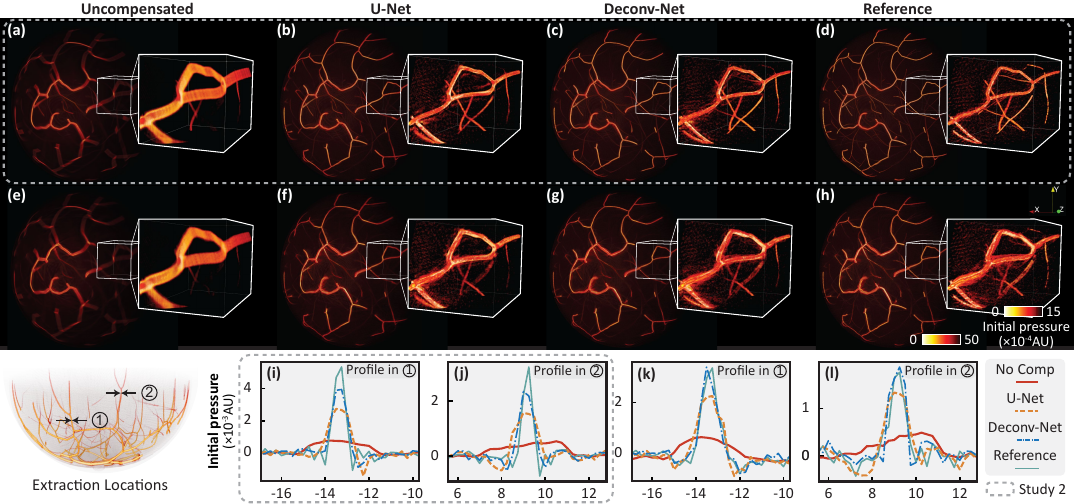} 
    \caption{Reconstructed images for a representative sample from the NBP-Homogeneous [(a)–(d)] and NBP-Heterogeneous [(e)–(h)] datasets. From left to right, panels (a, e) show uncompensated ${\mathbf{P}}_0^{\text{rect}}$; (b, f) show U-Net–compensated ${\mathbf{P}}_0^\text{U-Net}$; (c, g) show Deconv–Net–compensated ${\mathbf{P}}_0^{\text{Deconv-Net}}$; (d, h) show the reference $\mathbf {\mathbf{P}}_0^{\text{point}}$. Insets display zoomed-in regions for detailed visual comparison. Two vessel locations were selected for line profile analysis, as indicated in the 3D volume rendering on the bottom left. Line profiles extracted from reconstructed images displayed in panels (a-d) are shown in panels (i) and (j), and those from panels (e-h) are shown in panels (k) and (l). In the line profiles, uncompensated, U-Net-compensated, Deconv-Net-compensated, and reference results correspond to red solid, orange dashed, blue dashed, and turquoise solid lines, respectively. 
    }
    \label{fig:nbp_3d_rendering}
\end{figure*}

\section{Virtual Imaging Results} \label{sect:vit_results}
\subsection*{Study 1: Spatial Resolution Analysis} 
Figure~\ref{fig:study_1} 
analyzes spatial resolution
under the examined noise and SOS conditions using the Deterministic Spheres dataset. In panels (a--b), under the baseline conditions, ${\mathbf{P}}_0^{\text{rect}}$ shows increasing blurring for spheres positioned closer to the hemispherical measurement aperture, which has a radius of 85 mm. The blurring is anisotropic and occurs predominantly along the $y$-axis, which is approximately parallel to the aperture surface at those positions. In contrast, 
both ${\mathbf{P}}_0^\text{U-Net}$ and ${\mathbf{P}}_0^\text{Deconv-Net}$ exhibit 
mitigated blurring and closely resemble 
${\mathbf{P}}_0^{\text{point}}$.
In panel (c), the FWHM fitted from ${\mathbf{P}}_0^{\text{point}}$ ($\approx$0.505 mm) closely matches the true FWHM,
whereas ${\mathbf{P}}_0^{\text{rect}}$ shows 
resolution degradation with increasing $x$-coordinate. Both ${\mathbf{P}}_0^\text{U-Net}$ and ${\mathbf{P}}_0^\text{Deconv-Net}$ display relatively consistent resolution across all sphere positions, with FWHM values remaining close to 0.5 mm. These results demonstrate that the proposed SIR compensation framework effectively mitigated the resolution degradation observed in 
reconstructed images from 
SIR-uncompensated $\mathbf{P}^\text{rect}$.
Panel (d) in Fig.~\ref{fig:study_1} shows that both U-Net and Deconv-Net stably restored spatial resolution under the examined variations in noise and SOS. These results suggest that the proposed methods are robust to small perturbations in imaging conditions.



\subsection*{Study 2: Generalization to Object Complexity} 
\label{sect:study_2_results}
Panels (a) to (d) in Fig.~\ref{fig:nbp_3d_rendering} show representative reconstructed images corresponding to the NBP-Homogeneous dataset. The SIR-induced blurring in ${\mathbf{P}}_0^{\text{rect}}$ obscures vessel structures, particularly near the breast surface. Both compensation methods effectively recovered finer features, with ${\mathbf{P}}_0^\text{Deconv-Net}$ closely approximating the reference ${\mathbf{P}}_0^{\text{point}}$. This is further demonstrated by the line profiles shown in panels (i) and (j) of Fig.~\ref{fig:nbp_3d_rendering}.

\begin{table}[!t]
\centering
\scriptsize
\caption{Evaluation metrics for the NBP-Homogeneous dataset.}
\begin{tabular}{lcccc}
\toprule
\textbf{Metric} & \textbf{Depth} & \textbf{Uncompensated} & \textbf{U-Net} & \textbf{Deconv-Net} \\
\midrule
\multirow{2}{*}{RSE $\downarrow$} 
& Surface & 0.660 $\pm$ 0.061 & 0.405 $\pm$ 0.062 & \textbf{0.331 $\pm$ 0.048} \\
& Deeper  & 0.589 $\pm$ 0.086 & 0.510 $\pm$ 0.166 & \textbf{0.428 $\pm$ 0.144} \\
\midrule
\multirow{2}{*}{NCC $\uparrow$} 
& Surface & 0.582 $\pm$ 0.055 & 0.777 $\pm$ 0.037 & \textbf{0.822 $\pm$ 0.027} \\
& Deeper  & 0.640 $\pm$ 0.075 & 0.714 $\pm$ 0.108 & \textbf{0.768 $\pm$ 0.085} \\
\midrule
\multirow{2}{*}{DICE $\uparrow$} 
& Surface & 0.237 $\pm$ 0.048 & 0.758 $\pm$ 0.025 & \textbf{0.803 $\pm$ 0.011} \\
& Deeper  & 0.298 $\pm$ 0.052 & 0.700 $\pm$ 0.042 & \textbf{0.774 $\pm$ 0.022} \\
\bottomrule
\end{tabular}
\label{tab:study2_metrics}
\end{table}
Quantitative results (mean $\pm$ standard deviation) are summarized in Table~\ref{tab:study2_metrics}, where $\uparrow$ and $\downarrow$ indicate whether higher or lower values are preferable, respectively. For each metric and depth, the best result is boldfaced.
Both U-Net and Deconv-Net significantly improved RSE, NCC, and DICE scores over 
their uncompensated counterparts ($p < 0.05$ from Mann-Whitney U test), with Deconv-Net consistently outperforming U-Net across all metrics ($p < 0.05$).  
These qualitative and quantitative findings demonstrate that, 
although trained exclusively on the Stochastic Spheres dataset, the learned models generalize effectively 
to anatomically realistic and optically heterogeneous NBP data. 

\subsection*{Study 3: Generalization to Acoustic Heterogeneity}
\begin{table}[!t]
\centering
\scriptsize
\caption{Evaluation metrics for the NBP-Heterogeneous dataset.}
\begin{tabular}{lcccc}
\toprule
\textbf{Metric} & \textbf{Depth} & \textbf{Uncompensated} & \textbf{U-Net} & \textbf{Deconv-Net} \\
\midrule
\multirow{2}{*}{RSE $\downarrow$} 
& Surface & 0.596 $\pm$ 0.054 & 0.418 $\pm$ 0.065 & \textbf{0.382 $\pm$ 0.053} \\
& Deeper  & 0.586 $\pm$ 0.079 & 0.537 $\pm$ 0.134 & \textbf{0.469 $\pm$ 0.118} \\
\midrule
\multirow{2}{*}{NCC $\uparrow$} 
& Surface & 0.633 $\pm$ 0.044 & 0.774 $\pm$ 0.038 & \textbf{0.797 $\pm$ 0.029} \\
& Deeper  & 0.640 $\pm$ 0.066 & 0.699 $\pm$ 0.089 & \textbf{0.744 $\pm$ 0.072} \\
\midrule
\multirow{2}{*}{DICE $\uparrow$} 
& Surface & 0.237 $\pm$ 0.043 & 0.707 $\pm$ 0.045 & \textbf{0.738 $\pm$ 0.031} \\
& Deeper  & 0.249 $\pm$ 0.054 & 0.591 $\pm$ 0.087 & \textbf{0.648 $\pm$ 0.077} \\
\bottomrule
\end{tabular}
\vspace{-1mm}
\label{tab:study3_metrics}
\end{table}
The quantitative results of SIR compensation in the NBP-Heterogeneous dataset are summarized in Table~\ref{tab:study3_metrics}.
Despite the presence of acoustic variability not represented in the training data, images reconstructed from the compensated data with U-Net and Deconv-Net significantly outperformed those reconstructed from the uncompensated data across all metrics and depths (all $p < 0.05$). 
Deconv-Net consistently outperformed U-Net across all evaluations (all $p < 0.05$), which may be attributed to its incorporation of physics knowledge that constrains the solution space, reducing the complexity of the problem to be solved through learning. 
These findings indicate that the proposed methods retain robustness when applied to more realistic scenarios involving SOS heterogeneity, 
supporting their potential for practical deployment in experimental and clinical settings.

\normalcolor
\section{Application to Experimental Data} \label{sect:exp_study}
\subsection{Experiment Setup}
\normalcolor
\textit{In-vivo} data from the left and right breasts of a healthy volunteer were collected using the LOUISA-3D breast imaging system (TomoWave Laboratories, Houston, TX)~\cite{oraevsky2018full}.
The system's optical illumination and acoustic measurement subsystem geometries matched those used in 
the virtual imaging configuration described in Section~\ref{sect:virtual_imaging_system_configuration}.
The breast was illuminated at a wavelength of 797 nm. Each transducer, measuring 1.1 mm $\times$ 1.1 mm, recorded 1536 temporal samples at a sampling rate of 20 MHz.
To emulate measurements obtained using transducers measuring 5.5 mm $\times$ 1.1 mm, 
pressure data from every five vertically adjacent elements along the arc of the original 96-transducer array 
were averaged. 
While the topmost and bottommost two transducers were included in the averaging groups, their positions were not retained as the center locations of the emulated large transducers, resulting in 
a binned pressure data  $\mathbf{P}^\text{binned}$ for 92 transducer locations. 
The corresponding unbinned pressure data at these 92 positions were extracted and denoted as $\mathbf{P}^\text{unbinned}$. %

The proposed U-Net and Deconv-Net models were trained on the Stochastic Spheres dataset, with input data $\mathbf{P}^{\text{rect}}$ generated under configurations matching the experimental setup for the binned data, and target data $\mathbf{P}^{\text{ideal}}$ representing a point-like transducer equivalent.
The training procedure followed that described in Section~\ref {sect:model_training_and_image_recon}.
Once trained, the models were applied to the experimental data $\mathbf{P}^\text{binned}$ to produce compensated pressures $\mathbf{P}^\text{U-Net}$ and $\mathbf{P}^\text{Deconv-Net}$. 
From the pressure data $\mathbf{P}^\text{binned}$, $\mathbf{P}^\text{U-Net}$, $\mathbf{P}^\text{Deconv-Net}$, and $\mathbf{P}^\text{unbinned}$, image reconstruction was performed using the UBP method, yielding the corresponding images ${\mathbf{P}}_0^\text{binned}$, ${\mathbf{P}}_0^\text{U-Net}$, ${\mathbf{P}}_0^\text{Deconv-Net}$, and ${\mathbf{P}}_0^\text{unbinned}$, respectively. All images had dimensions $480\times 480\times 240$ voxels with an isotropic voxel size of 0.25 mm. A homogeneous SOS was assumed during reconstruction, with tuned values of 1.514 mm/$\mu$s for the left breast and 1.506 mm/$\mu$s for the right breast.

\begin{figure*}
    \centering
    \includegraphics[width=7.16in]{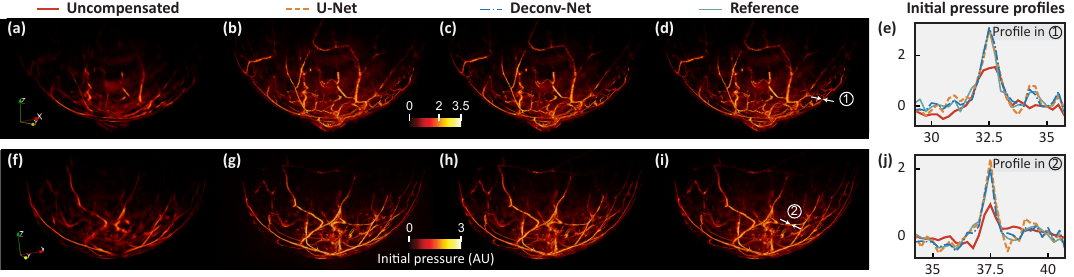} 
    \caption{Reconstructed images of \textit{in-vivo} data from the left [(a)-(d)] and right [(f)-(i)] breasts of a female subject. From left to right, the panels show results for binned data without SIR compensation [(a, f)], with SIR compensation using U-Net [panels (b, g)] and Deconv-Net [(c, h)], 
    and for the reference, unbinned data [(d, i)]. Panels (e) and (j) show line profiles of vessels extracted along the $y$-axis at positions indicated by white arrows in panels (d) and (i). 
    In the line profiles, the colors and line styles are consistent with those in Fig.~\ref{fig:nbp_3d_rendering}.}
    \label{fig:experimental_data}
    \vspace{-0.2cm}
\end{figure*}

For performance evaluation, the reconstructed images were visually compared, and line profiles were analyzed at vessel locations. Among these, ${\mathbf{P}}_0^\text{unbinned}$ was used as a reference, as it approximates the result expected from point-like transducers. Optimization-based reconstruction methods incorporating SIR were not employed due to their high computational cost, which arises from the large data size and the need to perform multiple reconstructions for parameter tuning. Moreover, because the transducer size was relatively small compared to the scanning radius, SIR effects were not apparent in ${\mathbf{P}}_0^\text{unbinned}$. 
\normalcolor
\subsection{Experimental Results}
\normalcolor
A visual comparison of the reconstructed images ${\mathbf{P}}_0^\text{binned}$, ${\mathbf{P}}_0^\text{U-Net}$, ${\mathbf{P}}_0^\text{Deconv-Net}$, and ${\mathbf{P}}_0^\text{unbinned}$ is shown in Fig.~\ref{fig:experimental_data}. In panels (a) and (f), ${\mathbf{P}}_0^\text{binned}$ exhibits significant SIR-induced artifacts, which obscure vessel structures. In contrast, ${\mathbf{P}}_0^\text{U-Net}$ in panels (b) and (g), and ${\mathbf{P}}_0^\text{Deconv-Net}$ in panels (c) and (h), more closely resemble the reference images shown in panels (d) and (i).
This is further illustrated by the line profiles in panels (e) and (j), where the width and values of the compensated results align well with the reference, showing 
significant improvement over 
the SIR-degraded ${\mathbf{P}}_0^\text{binned}$.

\section{Conclusion} 
\label{sect:conclusion}
\normalcolor
In this work, a learned data-space SIR compensation framework was proposed and validated.
Within the framework, a computationally efficient training data generation strategy tailored to the problem setting was introduced, leveraging analytic C-D forward simulations of stochastic sphere objects. Two SIR compensation models were explored: a purely data-driven U-Net and Deconv-Net that 
incorporates learned deconvolution filters motivated by the physics of the image formation. Systematic \textit{in-silico} studies validated the robustness of the proposed framework to variations in noise level, object complexity, and SOS heterogeneity. Evaluation using \textit{in-vivo} breast imaging data demonstrated the framework's generalizability to practical imaging conditions.

While the proposed framework effectively mitigates SIR artifacts, it is not intended to supplant optimization-based reconstruction methods that explicitly model the SIR for maximum accuracy. However, such methods typically require iterative computations of forward and adjoint models, along with careful tuning of associated parameters, including regularization parameters. The high computational cost hinders their practical usability, especially in clinical settings. In contrast, the proposed framework provides a fast and effective alternative means for suppressing SIR artifacts in time- or computationally-constrained applications.

Despite being trained solely on a synthetically generated Stochastic Spheres dataset, the proposed SIR compensation models demonstrated strong generalization to both anatomically realistic NBP dataset and \textit{in-vivo} breast imaging data. Each training sample consisted of 200 randomly placed spheres of varying sizes and intensities, simulated using a C-D forward model. When interpreted as a configuration of densely packed 1-voxel-diameter spheres, this setup is equivalent to a D-D model with spherical voxel bases~\cite{wang2010imaging}. The spatial distribution and pressure amplitudes in the training set were intentionally designed to fall within the typical range observed in actual breast imaging data. This signal-level alignment with the real-world conditions, despite the structural simplicity of the training objects, enabled the models to generalize effectively beyond their training domain.
The generalizability of the proposed models is further supported by the design choice of performing learning in the data domain rather than the image domain. Learning in the data domain reduces the risk of introducing false structures that are visually plausible but physically inaccurate. Image domain learning can inadvertently generate anatomically misleading features, particularly when trained on limited structural variability. However, errors introduced in the data domain are less likely to propagate into plausible but incorrect structures after reconstruction. This design choice provides a more interpretable and physically grounded correction mechanism, as supported by previous findings~\cite{cam2024learning,chen2025learning}.

In this work, the proposed framework was investigated under specific imaging system configurations. Nonetheless, the framework is not restricted to these settings and can be adapted to various imaging geometries, transducer sizes, and shapes. In the \textit{in-vivo} study, large transducer measurements were emulated by averaging data from adjacent small transducers. To more rigorously assess real-world performance, further experimental validation using systems equipped with physically large transducers is needed. As transducer sizes increase, SIR-induced information loss typically becomes more pronounced, highlighting the need for further investigation into the robustness of the proposed models across a range of transducer sizes. Additionally, since larger transducers may improve sensitivity and potentially extend imaging depth, future work should examine the trade-offs between transducer size, resolution, and imaging depth in the context of SIR compensation.


\bibliographystyle{IEEEtran}
\bibliography{main.bib}

\begin{thebibliography}{10}
\providecommand{\url}[1]{#1}
\csname url@samestyle\endcsname
\providecommand{\newblock}{\relax}
\providecommand{\bibinfo}[2]{#2}
\providecommand{\BIBentrySTDinterwordspacing}{\spaceskip=0pt\relax}
\providecommand{\BIBentryALTinterwordstretchfactor}{4}
\providecommand{\BIBentryALTinterwordspacing}{\spaceskip=\fontdimen2\font plus
\BIBentryALTinterwordstretchfactor\fontdimen3\font minus \fontdimen4\font\relax}
\providecommand{\BIBforeignlanguage}[2]{{%
\expandafter\ifx\csname l@#1\endcsname\relax
\typeout{** WARNING: IEEEtran.bst: No hyphenation pattern has been}%
\typeout{** loaded for the language `#1'. Using the pattern for}%
\typeout{** the default language instead.}%
\else
\language=\csname l@#1\endcsname
\fi
#2}}
\providecommand{\BIBdecl}{\relax}
\BIBdecl

\bibitem{wang2012photoacoustic}
L.~V. Wang and S.~Hu, ``Photoacoustic tomography: \textit{in vivo} imaging from organelles to organs,'' \emph{science}, vol. 335, no. 6075, pp. 1458--1462, 2012.

\bibitem{wang2017photoacoustic}
L.~Wang, \emph{Photoacoustic imaging and spectroscopy}.\hskip 1em plus 0.5em minus 0.4em\relax CRC press, 2017.

\bibitem{wang2009multiscale}
L.~V. Wang, ``Multiscale photoacoustic microscopy and computed tomography,'' \emph{Nature photonics}, vol.~3, no.~9, pp. 503--509, 2009.

\bibitem{wang2003noninvasive}
X.~Wang, Y.~Pang, G.~Ku, X.~Xie, G.~Stoica, and L.~V. Wang, ``Noninvasive laser-induced photoacoustic tomography for structural and functional \textit{in vivo} imaging of the brain,'' \emph{Nature biotechnology}, vol.~21, no.~7, pp. 803--806, 2003.

\bibitem{andreev2003detection}
V.~G. Andreev, A.~A. Karabutov, and A.~A. Oraevsky, ``Detection of ultrawide-band ultrasound pulses in optoacoustic tomography,'' \emph{IEEE transactions on ultrasonics, ferroelectrics, and frequency control}, vol.~50, no.~10, pp. 1383--1390, 2003.

\bibitem{kruger2010photoacoustic}
R.~A. Kruger, R.~B. Lam, D.~R. Reinecke, S.~P. Del~Rio, and R.~P. Doyle, ``Photoacoustic angiography of the breast,'' \emph{Medical physics}, vol.~37, no.~11, pp. 6096--6100, 2010.

\bibitem{ntziachristos2010molecular}
V.~Ntziachristos and D.~Razansky, ``Molecular imaging by means of multispectral optoacoustic tomography ({MSOT}),'' \emph{Chemical reviews}, vol. 110, no.~5, pp. 2783--2794, 2010.

\bibitem{oraevsky2018full}
A.~Oraevsky, R.~Su, H.~Nguyen, J.~Moore, Y.~Lou, S.~Bhadra, L.~Forte, M.~Anastasio, and W.~Yang, ``Full-view 3{D} imaging system for functional and anatomical screening of the breast,'' in \emph{Photons Plus Ultrasound: Imaging and Sensing 2018}, vol. 10494.\hskip 1em plus 0.5em minus 0.4em\relax SPIE, 2018, pp. 217--226.

\bibitem{oraevsky2000ultimate}
A.~A. Oraevsky and A.~A. Karabutov, ``Ultimate sensitivity of time-resolved optoacoustic detection,'' in \emph{Biomedical Optoacoustics}, vol. 3916.\hskip 1em plus 0.5em minus 0.4em\relax SPIE, 2000, pp. 228--239.

\bibitem{yang2007ring}
X.~Yang, M.-L. Li, and L.~V. Wang, ``Ring-based ultrasonic virtual point detector with applications to photoacoustic tomography,'' \emph{Applied physics letters}, vol.~90, no.~25, 2007.

\bibitem{wang2010imaging}
K.~Wang, S.~A. Ermilov, R.~Su, H.-P. Brecht, A.~A. Oraevsky, and M.~A. Anastasio, ``An imaging model incorporating ultrasonic transducer properties for three-dimensional optoacoustic tomography,'' \emph{IEEE transactions on medical imaging}, vol.~30, no.~2, pp. 203--214, 2010.

\bibitem{xu2005universal}
M.~Xu and L.~V. Wang, ``Universal back-projection algorithm for photoacoustic computed tomography,'' \emph{Physical Review E—Statistical, Nonlinear, and Soft Matter Physics}, vol.~71, no.~1, p. 016706, 2005.

\bibitem{anastasio2005half}
M.~A. Anastasio, J.~Zhang, X.~Pan, Y.~Zou, G.~Ku, and L.~V. Wang, ``Half-time image reconstruction in thermoacoustic tomography,'' \emph{IEEE transactions on medical imaging}, vol.~24, no.~2, pp. 199--210, 2005.

\bibitem{wang2011bookchapter}
K.~Wang and M.~A. Anastasio, \emph{Photoacoustic and thermoacoustic tomography: image formation principles}.\hskip 1em plus 0.5em minus 0.4em\relax New York, NY: Springer New York, 2011, pp. 781--815.

\bibitem{li2010model}
M.-L. Li, Y.-C. Tseng, and C.-C. Cheng, ``Model-based correction of finite aperture effect in photoacoustic tomography,'' \emph{Optics express}, vol.~18, no.~25, pp. 26\,285--26\,292, 2010.

\bibitem{rosenthal2011model}
A.~Rosenthal, V.~Ntziachristos, and D.~Razansky, ``Model-based optoacoustic inversion with arbitrary-shape detectors,'' \emph{Medical physics}, vol.~38, no.~7, pp. 4285--4295, 2011.

\bibitem{wang2013accelerating}
K.~Wang, C.~Huang, Y.-J. Kao, C.-Y. Chou, A.~A. Oraevsky, and M.~A. Anastasio, ``Accelerating image reconstruction in three-dimensional optoacoustic tomography on graphics processing units,'' \emph{Medical physics}, vol.~40, no.~2, p. 023301, 2013.

\bibitem{luo2023influences}
X.~Luo, J.~Jiang, H.~Wu, M.~Li, and B.~Wang, ``The influences of finite aperture size in photoacoustic computed tomography,'' \emph{Ultrasonics}, vol. 133, p. 107042, 2023.

\bibitem{lu2020full}
T.~Lu, Y.~Wang, J.~Li, J.~Prakash, F.~Gao, and V.~Ntziachristos, ``Full-frequency correction of spatial impulse response in back-projection scheme using space-variant filtering for optoacoustic mesoscopy,'' \emph{Photoacoustics}, vol.~19, p. 100193, 2020.

\bibitem{yanny2022deep}
K.~Yanny, K.~Monakhova, R.~W. Shuai, and L.~Waller, ``Deep learning for fast spatially varying deconvolution,'' \emph{Optica}, vol.~9, no.~1, pp. 96--99, 2022.

\bibitem{cam2024learning}
R.~M. Cam, U.~Villa, and M.~A. Anastasio, ``Learning a stable approximation of an existing but unknown inverse mapping: application to the half-time circular {R}adon transform,'' \emph{Inverse Problems}, vol.~40, no.~8, p. 085002, 2024.

\bibitem{munjal2024deep}
I.~Munjal and J.~Prakash, ``Deep-learning based deconvolution for correcting spatial impulse response of transducer in optoacoustic tomography,'' \emph{IEEE Transactions on Instrumentation and Measurement}, 2024.

\bibitem{yang2025compensating}
K.~Yang, S.~Park, H.-K. Huang, U.~Villa, and M.~A. Anastasio, ``Compensating for the spatial impulse response in 3{D} photoacoustic computed tomography using a learned data space restoration method,'' in \emph{Photons Plus Ultrasound: Imaging and Sensing 2025}, vol. 13319.\hskip 1em plus 0.5em minus 0.4em\relax SPIE, 2025, pp. 104--109.

\bibitem{stepanishen1971transient}
P.~R. Stepanishen, ``Transient radiation from pistons in an infinite planar baffle,'' \emph{The Journal of the Acoustical Society of America}, vol.~49, no.~5B, pp. 1629--1638, 1971.

\bibitem{jensen1999new}
J.~A. Jensen, ``A new calculation procedure for spatial impulse responses in ultrasound,'' \emph{The Journal of the Acoustical Society of America}, vol. 105, no.~6, pp. 3266--3274, 1999.

\bibitem{MITSUHASHI201421}
K.~Mitsuhashi, K.~Wang, and M.~A. Anastasio, ``Investigation of the far-field approximation for modeling a transducer's spatial impulse response in photoacoustic computed tomography,'' \emph{Photoacoustics}, vol.~2, no.~1, pp. 21--32, 2014.

\bibitem{wang2014discrete}
K.~Wang, R.~W. Schoonover, R.~Su, A.~Oraevsky, and M.~A. Anastasio, ``Discrete imaging models for three-dimensional optoacoustic tomography using radially symmetric expansion functions,'' \emph{IEEE transactions on medical imaging}, vol.~33, no.~5, pp. 1180--1193, 2014.

\bibitem{cciccek20163d}
{\"O}.~{\c{C}}i{\c{c}}ek, A.~Abdulkadir, S.~S. Lienkamp, T.~Brox, and O.~Ronneberger, ``3{D} {U}-{N}et: learning dense volumetric segmentation from sparse annotation,'' in \emph{Medical Image Computing and Computer-Assisted Intervention--MICCAI 2016: 19th International Conference, Athens, Greece, October 17-21, 2016, Proceedings, Part II 19}.\hskip 1em plus 0.5em minus 0.4em\relax Springer, 2016, pp. 424--432.

\bibitem{banerjee2024physics}
C.~Banerjee, K.~Nguyen, C.~Fookes, and K.~George, ``Physics-informed computer vision: A review and perspectives,'' \emph{ACM Computing Surveys}, vol.~57, no.~1, pp. 1--38, 2024.

\bibitem{li2024tabe}
A.~Li, G.~Yu, Z.~Xu, C.~Fan, X.~Li, and C.~Zheng, ``Tabe: Decoupling spatial and spectral processing with {T}aylor’s unfolding method in the beamspace domain for multi-channel speech enhancement,'' \emph{Information Fusion}, vol. 101, p. 101976, 2024.

\bibitem{park2023stochastic}
S.~Park, U.~Villa, F.~Li, R.~M. Cam, A.~A. Oraevsky, and M.~A. Anastasio, ``Stochastic three-dimensional numerical phantoms to enable computational studies in quantitative optoacoustic computed tomography of breast cancer,'' \emph{Journal of biomedical optics}, vol.~28, no.~6, pp. 066\,002--066\,002, 2023.

\bibitem{chen2025learning}
P.~Chen, S.~Park, R.~M. Cam, H.-K. Huang, A.~A. Oraevsky, U.~Villa, and M.~A. Anastasio, ``Learning a filtered backprojection reconstruction method for photoacoustic computed tomography with hemispherical measurement geometries,'' \emph{IEEE Transactions on Medical Imaging}, 2025.

\bibitem{american2018acr}
A.~C. of~Radiology, C.~J. D'Orsi \emph{et~al.}, \emph{ACR BI-RADS Atlas: Breast Imaging Reporting and Data System: 2013}.\hskip 1em plus 0.5em minus 0.4em\relax American college of Radiology, 2018.

\bibitem{fang2009monte}
Q.~Fang and D.~A. Boas, ``{M}onte {C}arlo simulation of photon migration in 3{D} turbid media accelerated by graphics processing units,'' \emph{Optics express}, vol.~17, no.~22, pp. 20\,178--20\,190, 2009.

\bibitem{tabei2002k}
M.~Tabei, T.~D. Mast, and R.~C. Waag, ``A \textit{k}-space method for coupled first-order acoustic propagation equations,'' \emph{The Journal of the Acoustical Society of America}, vol. 111, no.~1, pp. 53--63, 2002.

\bibitem{treeby2010photoacoustic}
B.~E. Treeby, E.~Z. Zhang, and B.~T. Cox, ``Photoacoustic tomography in absorbing acoustic media using time reversal,'' \emph{Inverse Problems}, vol.~26, no.~11, p. 115003, 2010.

\bibitem{jax2018github}
\BIBentryALTinterwordspacing
J.~Bradbury, R.~Frostig, P.~Hawkins, M.~J. Johnson, C.~Leary, D.~Maclaurin, G.~Necula, A.~Paszke, J.~Vander{P}las, S.~Wanderman-{M}ilne, and Q.~Zhang, ``{JAX}: composable transformations of {P}ython+{N}um{P}y programs,'' 2018. [Online]. Available: \url{http://github.com/jax-ml/jax}
\BIBentrySTDinterwordspacing

\bibitem{stanziola2023j}
A.~Stanziola, S.~R. Arridge, B.~T. Cox, and B.~E. Treeby, ``j-{W}ave: An open-source differentiable wave simulator,'' \emph{SoftwareX}, vol.~22, p. 101338, 2023.

\bibitem{wise2019representing}
E.~S. Wise, B.~Cox, J.~Jaros, and B.~E. Treeby, ``Representing arbitrary acoustic source and sensor distributions in {F}ourier collocation methods,'' \emph{The Journal of the Acoustical Society of America}, vol. 146, no.~1, pp. 278--288, 2019.

\bibitem{frangi1998multiscale}
A.~F. Frangi, W.~J. Niessen, K.~L. Vincken, and M.~A. Viergever, ``Multiscale vessel enhancement filtering,'' in \emph{Medical image computing and computer-assisted intervention—MICCAI’98: first international conference cambridge, MA, USA, october 11--13, 1998 proceedings 1}.\hskip 1em plus 0.5em minus 0.4em\relax Springer, 1998, pp. 130--137.

\bibitem{zack1977automatic}
G.~W. Zack, W.~E. Rogers, and S.~A. Latt, ``Automatic measurement of sister chromatid exchange frequency,'' \emph{Journal of Histochemistry \& Cytochemistry}, vol.~25, no.~7, pp. 741--753, July 1977.

\bibitem{mann1947test}
H.~B. Mann and D.~R. Whitney, ``On a test of whether one of two random variables is stochastically larger than the other,'' \emph{The annals of mathematical statistics}, pp. 50--60, 1947.

\end{thebibliography}




\end{document}